\documentclass[letterpaper, 10 pt, conference]{ieeeconf}  

\IEEEoverridecommandlockouts                              

\usepackage{graphicx} 
\usepackage{amsmath} 
\usepackage{amssymb}  
\usepackage{hyperref}
\usepackage{booktabs}
\usepackage{multirow}
\usepackage{tabularx}

\renewcommand{\thefootnote}{}

\begin{document}

\title{\LARGE \bf
InstruGen: Automatic Instruction Generation for Vision-and-Language Navigation Via Large Multimodal Models
}

\author{Yu Yan\textsuperscript{*}, Rongtao Xu\textsuperscript{*}, Jiazhao Zhang, Peiyang Li, Xiaodan Liang, Jianqin Yin$^{\dag}$}

\maketitle

\renewcommand{\thefootnote}{}
\footnotetext{
\textsuperscript{*} Yu Yan and Rongtao Xu contributed equally.

$^{\dag}$Jianqin Yin is the corresponding author (jqyin@bupt.edu.cn).

Yu Yan and Peiyang Li are with the Beijing University of Posts and Telecommunications, China.
Rongtao Xu is with the Institute of Automation, Chinese Academy of Sciences, China. 
Jiazhao Zhang is with Peking University, China. Xiaodan Liang is with Sun Yat-Sen University, China. 

}

\begin{abstract}

Recent research on Vision-and-Language Navigation (VLN) indicates that agents suffer from poor generalization in unseen environments due to the lack of realistic training environments and high-quality path-instruction pairs. Most existing methods for constructing realistic navigation scenes have high costs, and the extension of instructions mainly relies on predefined templates or rules, lacking adaptability. To alleviate the issue, we propose \textit{InstruGen}, a VLN path-instruction pairs generation paradigm. Specifically, we use YouTube house tour videos as realistic navigation scenes and leverage the powerful visual understanding and generation abilities of large multimodal models (LMMs) to automatically generate diverse and high-quality VLN path-instruction pairs. Our method generates navigation instructions with different granularities and achieves fine-grained alignment between instructions and visual observations, which was difficult to achieve with previous methods. Additionally, we design a multi-stage verification mechanism to reduce hallucinations and inconsistency of LMMs. Experimental results demonstrate that agents trained with path-instruction pairs generated by \textit{InstruGen} achieves state-of-the-art performance on the R2R and RxR benchmarks, particularly in unseen environments. Code is available at \small{\url{https://github.com/yanyu0526/InstruGen}}

\end{abstract}

\section{INTRODUCTION}

Vision-and-Language Navigation (VLN) requires agents to navigate in the visual environment based on natural language instructions. Diverse and realistic visual scenes, along with accurate and detailed instructions, are essential for the agent to navigate \cite{chen2022learning, kamath2023new, zhang2024navid}. Since the release of datasets like Room-to-Room (R2R) \cite{anderson2018vision}, many VLN datasets have been introduced. However, the lack of high-quality path-instruction pairs and the limited realistic training environments lead to overfitting to seen environments and poor generalization to unseen ones \cite{zhang2020diagnosing}. 

Data-centric approaches are an effective method for alleviating overfitting. Some studies have attempted to expand data in simulators like Matterport3D \cite{MP3d} by dropouting environmental features \cite{EnvDrop}, changing image styles \cite{ma2019regretful}, or using speaker models to generate path-instruction pairs \cite{EnvDrop, fried2018speaker}. To further enhance the diversity of VLN training resources, VLNbert \cite{Improving} and Airbert \cite{guhur2021airbert} have employed web image captions to generate path-instruction pairs. However, simply concatenating web images cannot fully replicate the authentic navigation experience, thus limiting the agent's navigation performance. Consequently, \cite{Kumar2019LearningNS,chang2020semantic,lin2023learning} train agents using YouTube house tour videos to provide a more realistic navigation experience. Specifically, \cite{lin2023learning} develops the YouTube-VLN dataset, featuring VLN-like path-instruction pairs for in-domain pre-training. However, these methods rely on predefined templates or manual rules to generate navigation instructions, which not only have a fixed format but make it challenging to achieve fine-grained alignment between instructions and visual observations, resulting in inaccurate instructions and limiting the temporal grounding ability of VLN agents \cite{SMNA}.

\begin{figure}[!t]
\centering 
\includegraphics[width=0.99\linewidth]{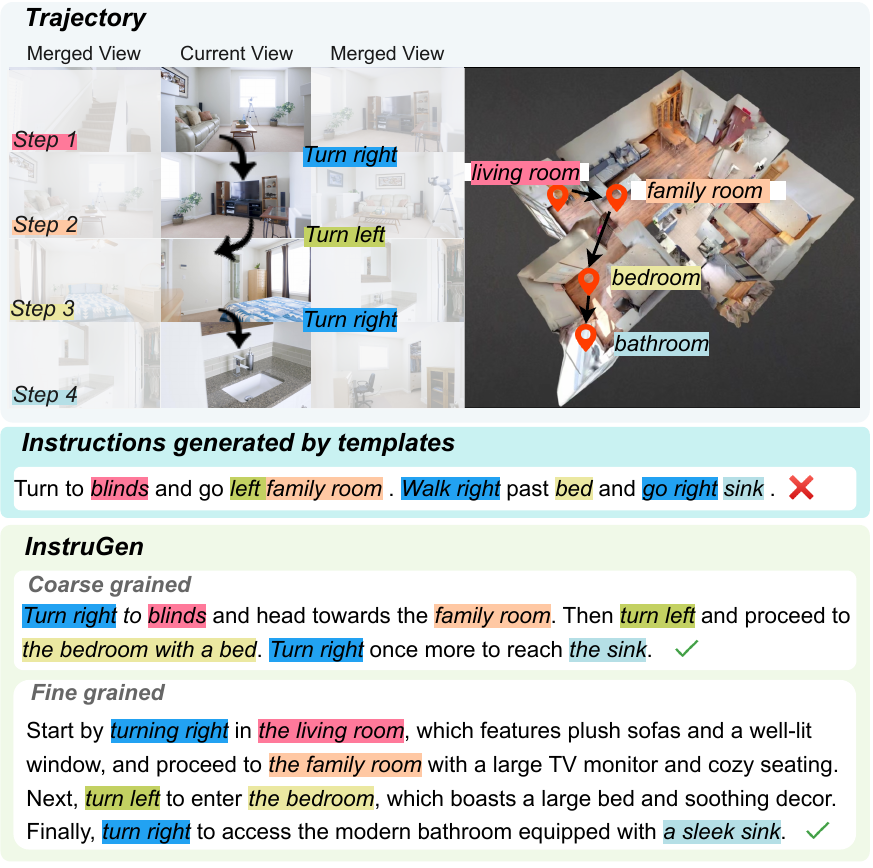}
\caption{Generate navigation instructions for trajectories sampled from YouTube videos. Instructions generated using R2R templates often deviate from the actual trajectory. The instructions generated by \textit{InstruGen} are more precise and express environmental information in detail.}
\label{Example}
\end{figure}

In recent years, large multimodal models (LMMs) have developed rapidly, especially with the advent of ChatGPT-4V \cite{yang2023dawn}, which can process both image and text inputs and generate text outputs. In light of this, we propose a VLN path-instruction pairs generation paradigm called \textit{InstruGen}, which completely discards the limitations of rigid templates and limited visual scenes for data augmentation and achieves infinite expansion of VLN agent training resources. Specifically, inspired by \cite{lin2023learning}, we use YouTube house tour videos as realistic navigation scenes and sample diverse navigation trajectories. Leveraging the powerful visual understanding and generation abilities of the ChatGPT-4V model, we analyze objects, scenes, and their layouts at each trajectory node. Then, we integrate the visual information to form a global understanding of the trajectory and automatically generate high-quality instructions that match the trajectory. It allows us to generate high-quality instructions that match the trajectories automatically (see Fig. \ref{Example}). Although ChatGPT-4V can generate highly credible instructions, it can still suffer from hallucinations, such as inaccurate action sequences, erroneous trajectory node descriptions, and temporal and spatial logical inconsistencies in the instructions. Therefore, we design a multi-stage verification mechanism to verify whether the action and node descriptions are consistent with the input trajectory, sentence completeness, etc., effectively reducing hallucinations of ChatGPT-4V.

Our empirical studies show that \textit{InstruGen} can generate precise and detailed navigation instructions for any trajectory. These instructions are easy to understand and can be adjusted according to the needs for the granularity of navigation instructions. More importantly, \textit{InstruGen} can accurately interpret complex visual scenes, ensuring logical consistency and alignment between trajectories and instructions. 

Our contributions are summarized as the following:
\begin{itemize}
    \item we propose the path-instruction pairs generation paradigm called \textit{InstruGen}, utilizing ChatGPT-4V to construct a high-quality instruction generation system that can accurately match navigation paths, achieving large-scale expansion of VLN training resources and proposing corresponding datasets.
    \item We design a multi-stage verification mechanism that significantly mitigates hallucinations of ChatGPT-4V, ensuring the accuracy and consistency of the path-instruction pairs.
    \item Our agents demonstrate outstanding navigation and generalization abilities on the R2R and RxR benchmarks, achieving the best performance. It further highlights the importance of high-quality navigation training resources in enhancing agent performance and validates the effectiveness of our approach.
\end{itemize}

\section{RELATED WORK}

\subsection{Vision-and-Language Navigation}

Interacting with agents using natural language is a long-standing goal. Towards this goal, Vision-and-Language Navigation (VLN) \cite{chen2019touchdown, thomason2020vision} has been proposed, covering various tasks from strictly following initial instructions \cite{anderson2018vision, yan2019cross, RxR, he2021landmark} to actively exploring the environment and interacting with objects \cite{qi2020reverie, padmakumar2022teach}. Nonetheless, many challenges exist in VLN tasks, including understanding and alignment of information from different modalities, reasoning strategies for navigation, the issue of data scarcity, and models' generalization abilities in unseen environments \cite{gu-etal-2022-vision}. To address these challenges, researchers have devised numerous strategies, encompassing representation learning \cite{kim2021ndh, Improving, guhur2021airbert}, action strategy learning \cite{anderson2018vision, he2021landmark}, data-centric \cite{EnvDrop, fried2018speaker, zhu2020multimodal}, and prior exploration \cite{parvaneh2020counterfactual, wang2019reinforced}. These approaches aim to improve models' comprehension of multimodal information, facilitate informed decision-making, mitigate the impact of limited data, and enhance generalization to unseen environments. We have thoroughly investigated overcoming the challenges of data scarcity and generalization in VLN tasks.

\subsection{Data-centric VLN approaches}

Data-centric approaches enhance VLN training resources by utilizing existing data or generating synthetic data. For Environment Augmentation, novel environments are crafted by randomly masking visual features across viewpoints \cite{EnvDrop}, segmenting and recombining house scenes \cite{liu2021vision}, or altering styles and objects \cite{li2022envedit}. Although the scenes are greatly enriched, they are still based on limited environments, and generating corresponding instructions for distinct visual features in different environments remains challenging. For Path-Instruction Augmentation, Fried \cite{fried2018speaker} demonstrates that training a Speaker model to generate instructions based on specific navigation paths significantly improves agents' performance. ProbeES \cite{Prompt} and AutoVLN \cite{chen2022learning} enrich navigation instructions through self-discovery in simulated environments. VLNbert \cite{Improving} and Airbert \cite{guhur2021airbert} are not limited to simulator environments and utilize web images and captions for pre-training data. Lily \cite{lin2023learning} leverages house tour videos from YouTube to provide a pre-training dataset based on real-world experiences, enhancing data authenticity and applicability. Marky \cite{kamath2023new}, a generator trained with text aligned to visual landmarks, achieves near-human quality on R2R-style paths in unseen environments. 

Although the above work has made significant contributions, there are still issues with limited real-world navigation environments and reliance on fixed templates or rules for generating instructions. In contrast, we utilize YouTube house tour videos as navigation scenes. These videos are abundant, cost-effective, and offer diverse and realistic environments. In addition, we leverage LMMs to construct an automated, high-quality instruction generation system that matches navigation paths accurately.

\begin{figure*}[!t]
\centering
\includegraphics[width=0.99\linewidth]{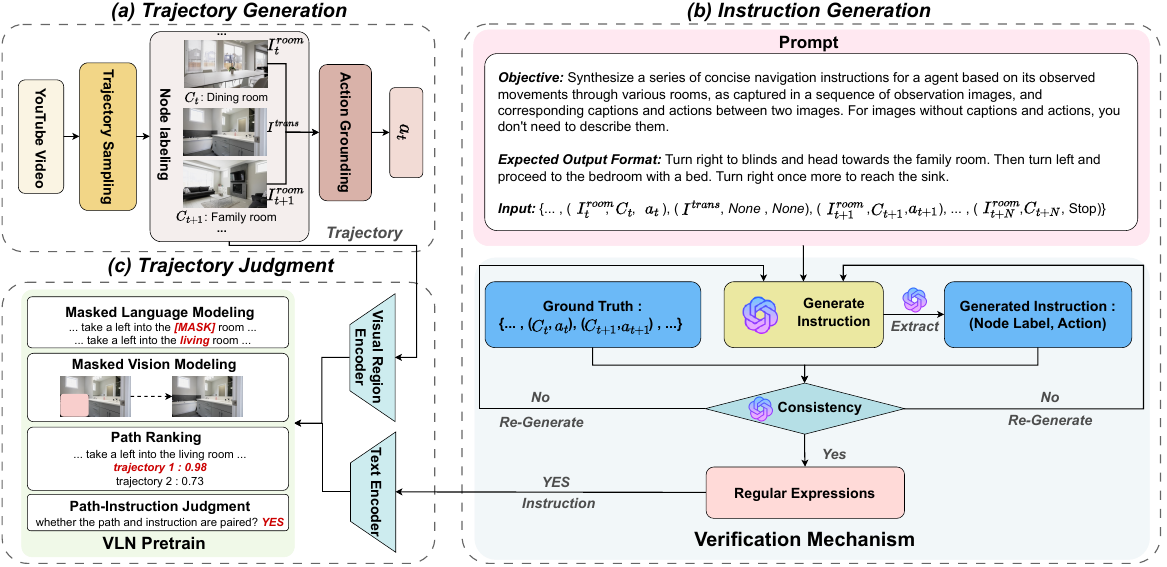}
\caption{\textit{InstruGen} consists of the following stages: (a) Trajectory Generation, (b) Instruction Generation, and (c) Trajectory Judgment.}
\label{framework}
\end{figure*}

\section{METHOD}

\textit{InstruGen} is an automated pipeline comprising three stages(see Fig. \ref{framework}). (a) Trajectory Generation: we sample diverse trajectories and annotate the nodes and actions between nodes, representing the trajectory as a series of triplets (image, node label, action). (b) Instruction Generation: through prompt engineering, we guide ChatGPT-4V to generate customized instructions based on specific needs and design a multi-stage verification mechanism to ensure the quality of the instructions. (c) Trajectory Judgment: we design various pretext tasks to ensure the reasonableness of the path-instruction pairs.

\subsection{Trajectory Generation}

\textbf{Trajectory Sampling.} We use 3879 YouTube house tour videos as realistic indoor navigation scenes. To convert these videos into discrete navigation trajectories similar to R2R, we sample trajectories following \cite{lin2023learning}. Each trajectory consists of room nodes and transition nodes, i.e., $ I=\left \{ ...,I^{room}_t,I^{trans},I^{room}_{t+1},... \right \} $, where $I^{room}_t$ denotes the image of the room node at time \textit{t}, $I^{trans}$ denotes the image of the transition node. 

\textbf{Node Labeling.} We found that directly inputting trajectory images into ChatGPT-4V may cause the model to experience hallucinations when interpreting the visual information, such as misidentifying the room type of nodes or adding objects not present in the image. Therefore, we utilize CLIP \cite{radford2021learning} to annotate each node with its room type and critical objects in the image, represented as 'Object with Room.' Transitional nodes generally do not contain valuable information, so we do not annotate them. Ultimately, we obtain trajectories with node label information, denote as  $ \left \{ ...,I^{room}_t,C_t,I^{trans},I^{room}_{t+1},C_{t+1},... \right \} $, $C_t$ denotes the label of the room node at time \textit{t}, which guide ChatGPT-4V to understand the information of navigation nodes accurately. It ensures the accuracy of the trajectories and further enhances the model's understanding of the navigation environment.

\textbf{Action Grounding.} It is necessary to annotate the transition actions between nodes when ChatGPT-4V directly interprets the transitions between trajectory nodes to avoid logical confusion. Given the high cost of manually annotating video data, we introduce the action inverse model \cite{Kumar2019LearningNS} to estimate the specific actions performed during the transitions between room nodes in the trajectory, denote as $\left \{ ...,I^{room}_t,a_t,I^{room}_{t+1},... \right \}$.

\subsection{Instruction Generation}

\textbf{Task Proposal.} We utilize ChatGPT-4V as LMMs for instruction generation, which can be replaced with the optimal model at any time. ChatGPT-4V is trained on diverse datasets, endowing it with rich domain knowledge and abilities. The autoregressive instruction refinement also enhances the model's responsiveness to given prompts. It allows us to activate its expertise and domain-specific abilities by defining precise task requirements within the input prompts. Consequently, we devise the instruction generation task by setting specific prompts (see Fig. \ref{framework}(b)). 

The accuracy of navigation instructions depends on correctly capturing the visual information of trajectory nodes and the transition actions between them. We organize the trajectory sequences, the node label and action sequences generated during the section A, into a series of triplets as input for ChatGPT-4V, formatted as follows: $\{ ...,(I^{room}_t,C_t,a_t), (I^{trans},None,None), (I^{room}_{t+1},C_{t+1}, $ $a_{t+1}),...,(I^{room}_{t+N},C_{t+N},Stop)\} $. To ensure the instructions are more realistic, we manually set the action of the final node in the sequence to “\textit{Stop}.” To further prevent potential hallucinations when ChatGPT-4V interprets the visual information of nodes, we explicitly indicate in the task prompt that no additional descriptions of the node's visual information are needed beyond the provided labels. Additionally, we provide a prompt that includes an example of the expected output format to enhance the model's guidance. By specifying the granularity requirements (concise or detailed) in the task definition and altering the output format (whether to include environmental descriptions and key objects), we can effectively guide ChatGPT-4V to generate navigation instructions with varying levels of detail.

\begin{figure}[!t]
\centering 
\includegraphics[width=\linewidth,height=0.87\linewidth]{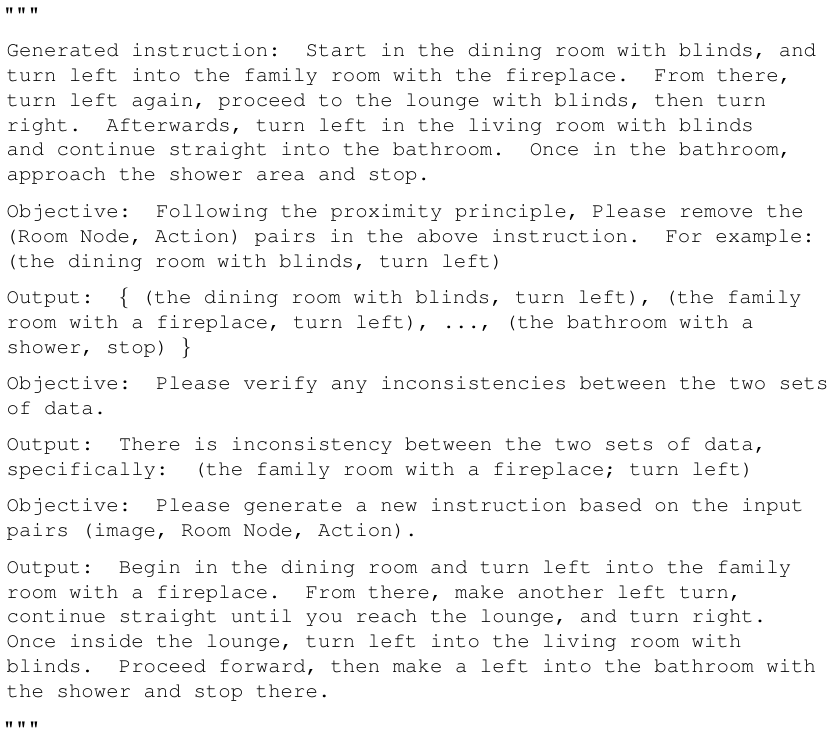}
\caption{Prompt for the Consistency module}
\label{consistency}
\end{figure}

\textbf{Verification Mechanism.} Our validation mechanism consists of two modules: Consistency and Regular Expression (see Fig. \ref{framework}(b)). The consistency module uses the trajectory from the Input of the prompt as the Ground Truth. First, we design prompts (see Fig. \ref{consistency}) to guide ChatGPT-4V in extracting (node label, action) pairs from the generated instructions based on proximity. For example, for the instruction \textit{"Start from the dining room, turn left into the family room, then go straight into the living room,"} we extract pairs such as \textit{"(dining room, turn left)"} and \textit{"(family room, go straight)."} These extracted (node label, action) pairs are then compared with the Ground Truth for consistency. If mismatches are detected, the system automatically triggers a regeneration process to ensure that the output instructions precisely correspond to the given navigation trajectory. Additionally, ChatGPT-4V may introduce unintended 'special characters' or portions of the 'prompt' during the instruction generation process. To resolve these issues, we implement regular expression-based rules to make subtle adjustments to the instructions, eliminating any logical inconsistencies or linguistic errors.  

\subsection{Trajectory Judgment}

\textbf{Model Architecture.} We draw inspiration from the studies of Lily \cite{lin2023learning} and Airbert \cite{guhur2021airbert}, adopting an architecture similar to ViLBERT \cite{lu2019vilbert} as the backbone of our model to extract visual and textual features (see Fig. \ref{framework}(c)). Given the trajectory $I$ and the corresponding instruction $W$, where $I=\left\{ i^k\right \}_{k=1}^K $ and $K$ represents the number of nodes. The model encodes each node $i^k$ into regional features $\left[ i_1^k,...,i_n^k \right ]$. Respectively, the visual and text inputs to our model are :

{
\begin{equation}
\label{eq2}
X_I=\left [ [IMG],i_1^1,...,i_n^1,...,[IMG],i_1^K,...,i_n^K \right ]
\end{equation}
\begin{equation}
\label{eq2}
X_W=\left [[CLS],w_1,...,w_T,[SEP]\right]
\end{equation}
}

Where $\left[CLS\right ]$ and $\left[SEP\right ]$ are unique tokens added to the text, the $\left[IMG\right]$ token is used to separate image region features taken at different locations.

\textbf{Pretext Tasks for Pre-training.} Based on our ViLBERT-like \cite{lu2019vilbert} model architecture, we design three pretext tasks: Masked Language Modeling (MLM), Masked Vision Modeling (MVM), and Path-Instruction Judgment (PIJ). The MLM and MVM tasks randomly replace words or image regions with [MASK] labels, and the model predicts these masked contents in multimodal contexts. The PIJ task evaluates whether paths and instructions are paired, so We consider the trajectories generated by ChatGPT-4V as positive samples and create negative samples by randomly shuffling the order of trajectory nodes to train the model. The training model evaluates whether the path and instructions are paired. Additionally, inspired by \cite{EnvDrop}, the agent generates multiple candidate paths, and the model is trained to select the best path. Therefore, we design the Path Ranking (PR) task, where the model scores each candidate path based on the extracted features, assessing how well it matches the instructions. This allows the agent to choose the highest-scoring path as the final navigation route.

\section{EXPERIMENTS}

\subsection{Experimental Setup}

\textbf{Dataset.} We conduct our experiments on two VLN benchmarks, i.e., Room-to-Room (R2R) \cite{anderson2018vision} and Room-Across-Room (RxR) \cite{RxR}. The R2R benchmark covers 90 buildings and provides over 21,000 fine-grained natural language instructions. RxR is an extension of R2R, which offers large-scale multilingual instructions while increasing the number and complexity of paths. The instructions are more detailed and contain richer environmental information, making navigation more challenging. Through experiments on the R2R and RxR, we can comprehensively evaluate the navigation abilities of agents in different linguistic and complex environmental settings.

\textbf{Implement details.} We pre-train and fine-tune the model using constructed YouTube path-instruction pairs. We use 3679 videos for training in seen environments and 200 videos reserved for validation in unseen environments. We use 4 NVIDIA 4090 GPUs with the batch size of 4 (1 batch per GPU). We conducted 50 epochs during the pre-training process and trained for approximately one week. We conducted 30 epochs on R2R and R2R \# during the fine-tuning process and trained for approximately three days. On R2R*, we conducted 50 epochs and trained for approximately five days. 

\subsection{Pre-Training with InstruGen Data}

\textbf{The impact of verification mechanisms on model performance and generated instruction quality.} We conduct experiments to validate the significant effect of the verification mechanism on model performance. Table \ref{verification ablation} shows the experimental results of different verification mechanism modules on R2R Val Unseen datasets. Comparing \#1 with other rows, we can conclude that using the verification mechanism results in significantly better model performance than not using it. The comparison \#1 and \#2, \#2 and \#3, indicates that the consistency module has strong self-correction capabilities. Comparing \#3 and \#4, we find that the model's performance with only the verification mechanism differs slightly from the model manually inspected, suggesting that the accuracy of instructions after applying the verification mechanism is already close to human-level performance.

\begin{figure}[!t]
\centering 
\includegraphics[width=0.99\linewidth]{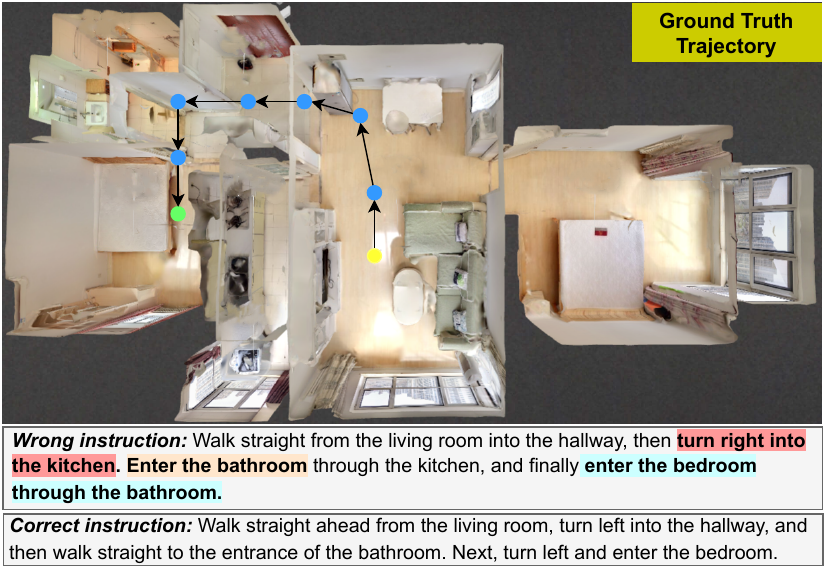}
\caption{Impact of verification mechanism on instruction quality. Yellow and green circles represent the starting and target positions, respectively.}
\label{verification visiual}
\end{figure}

Furthermore, Fig. \ref{verification visiual} visualizes the impact of the verification mechanism on instruction quality. Without the verification mechanism, the generated instructions often contain logical errors and inconsistencies, leading to navigation paths deviating from the target. For instance, turning right into the kitchen is impossible, and the agent does not enter the bathroom, which is a step that is inconsistent with the actual path. However, with the verification mechanism, the generated instructions are more accurate and consistent. These results demonstrate that our verification mechanism improves data quality and reduces model errors, enhancing the agent's navigation performance in complex environments.

\begin{table}[!t]
{
\resizebox{0.98\linewidth}{!}{
    \begin{tabular}{c|cc|c|cccc}
    \hline
    \multirow{2}{*}{\#} & \multicolumn{2}{c|}{Verification Mechanism}                                  & \multirow{2}{*}{\begin{tabular}[c]{@{}c@{}}Manual\\ Review\end{tabular}} & \multicolumn{4}{c}{Val-Unseen} \\ \cline{2-3} \cline{5-8} 
                    & Consistency & \begin{tabular}[c]{@{}c@{}}Regular\\ Expressions\end{tabular} &                                                                          & TL    & NE ↓   & SR  ↑   & SPL ↑    \\ \hline
1                   &  -           & -                    & -              & 10.15      & 3.47      & 67.42       & 0.62      \\ 
2                   &  \checkmark           &  -                   & -              & 9.98      & 3.27     & 69.32       & 0.64      \\ 
3                   &  \checkmark           &  \checkmark                   & -              & 9.76      & 3.15      & 70.9       & 0.65      \\ \hline
4                   & \checkmark            & \checkmark                    & \checkmark              & 9.80  & \textbf{3.13}  & \textbf{71.10}  & \textbf{0.66}  \\ \hline
    \end{tabular}
}
}
\centering
\caption{Ablation study of the verification mechanism on R2R.}
\label{verification ablation}
\end{table}

\begin{figure}[!t]
\centering 
\includegraphics[width=0.99\linewidth]{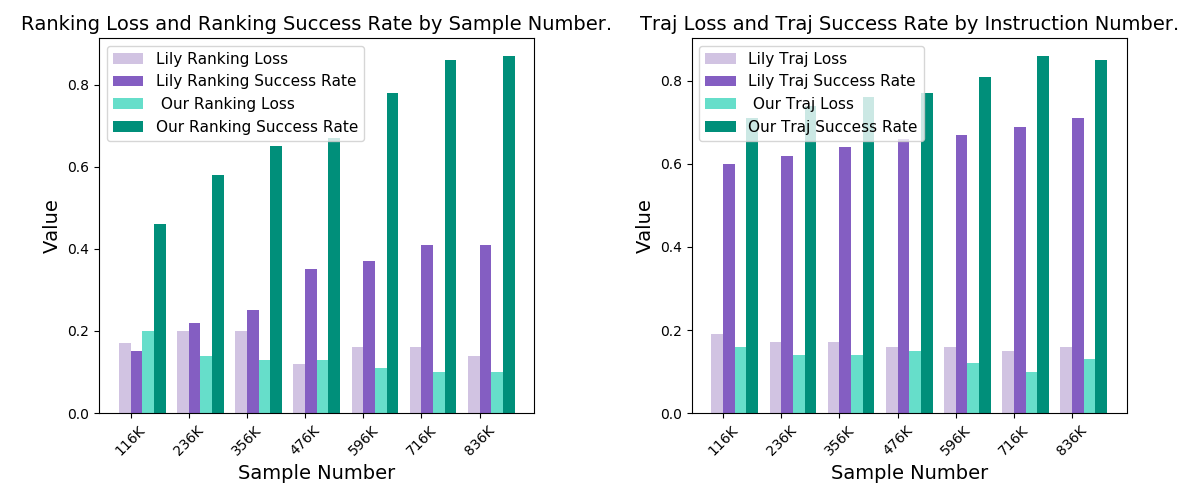}
\caption{The impact of data quality on loss and accuracy during the pre-training process.}
\label{loss and accuracy}
\end{figure}

\textbf{The impact of data quality on loss and accuracy during pre-training.} We believe data quality significantly affects loss and accuracy during pre-training. High-quality instructions, compared to low-quality ones, help the model effectively learn valuable features, reduce the loss value, and enable faster convergence. In our experiments, we compare the pre-training performance of our model with Lily under different data quality conditions (see Fig. \ref{loss and accuracy}). Ranking refers to ranking trajectories during training, while Traj refers to determining whether the path is reasonable. The results show that, given the same sampled data, our method outperforms Lily in terms of "Ranking Loss" and "Traj Loss," demonstrating better training stability and convergence. Our process also surpasses Lily in "Ranking Success Rate" and "Traj Success Rate," with a increase in accuracy, especially in the later stages of training. These results further validate the importance and effectiveness of high-quality data in enhancing model performance during pre-training.

\begin{table}[!t]
{
\resizebox{1.0\linewidth}{!}{
    \begin{tabular}{c|cc|ccc|c|cccc}
\hline
\multirow{2}{*}{\#} & \multicolumn{2}{c|}{Pretraining}                                                                                   & \multicolumn{3}{c|}{Finetuing} & Testing & \multicolumn{4}{c}{Val Unseen} \\ \cline{2-11} 
                    & \begin{tabular}[c]{@{}c@{}}Coarse\\ grained\end{tabular} & \begin{tabular}[c]{@{}c@{}}Fine\\ grained\end{tabular} & R2R     & R2R\#     & R2R*    & R2R     & TL    & NE ↓   & SR↑    & SPL ↑   \\ \hline
1                   &  -                                                                     & \checkmark      & \checkmark  & -         &  -        & \checkmark   & 9.97     & 3.88      & 63.97      & 0.59       \\ 
2                   &  -                                                                     & \checkmark     & -        & \checkmark         & -  & \checkmark         & 10.12      &  3.43     & 67.34      & 0.62       \\ 
3                    &  -                                                                     & \checkmark     &  -       & -  & \checkmark         & \checkmark     &  10.02    & 3.35    & 68.13      & 0.61       \\  \hline
4                   & \checkmark                  & -                                   & \checkmark &-          &  -        & \checkmark    & 9.64      & 3.28      & 69.93      & 0.64       \\ 
5                   & \checkmark       &  -                                                                  &   -      &   \checkmark      & -  & \checkmark        & 9.77      &  3.14     & 70.8      & 0.65 \\
6                  & \checkmark          & -          & -        & -  & \checkmark      & \checkmark     & 9.80      & \textbf{3.13}      & \textbf{71.10}      & \textbf{0.66}       \\ \hline
\end{tabular}

}
}
\centering
\caption{The model's performance with instructions of different granularities on R2R.}
\label{style}
\end{table}

\textbf{The impact of instructions of different granularities on model performance.} Different granularities of instructions, such as Concise or detailed navigation description, affect the agent's understanding and execution abilities. In Table \ref{style}, R2R\# refers to the instruction to rewrite the R2R trajectory using \textit{InstruGen}. R2R* is a mixed instruction set of R2R\# and R2R. Comparing \#1-3 and \#4-6, we found that the model pre-trained with coarse-grained instructions performed better than that trained with fine-grained instructions. Comparing \#1 and \#2, \#4 and \#5, we found that precise instructions have a more significant impact on improving the navigation ability of the agent. Comparing \#2 and \#3, \#5 and \#6 found that the model fine-tuned with R2R* performed better due to seeing more instructions, indicating that rich instructions significantly impact the model's generalization ability.

\begin{figure}[!t]
\centering 
\includegraphics[width=\linewidth]{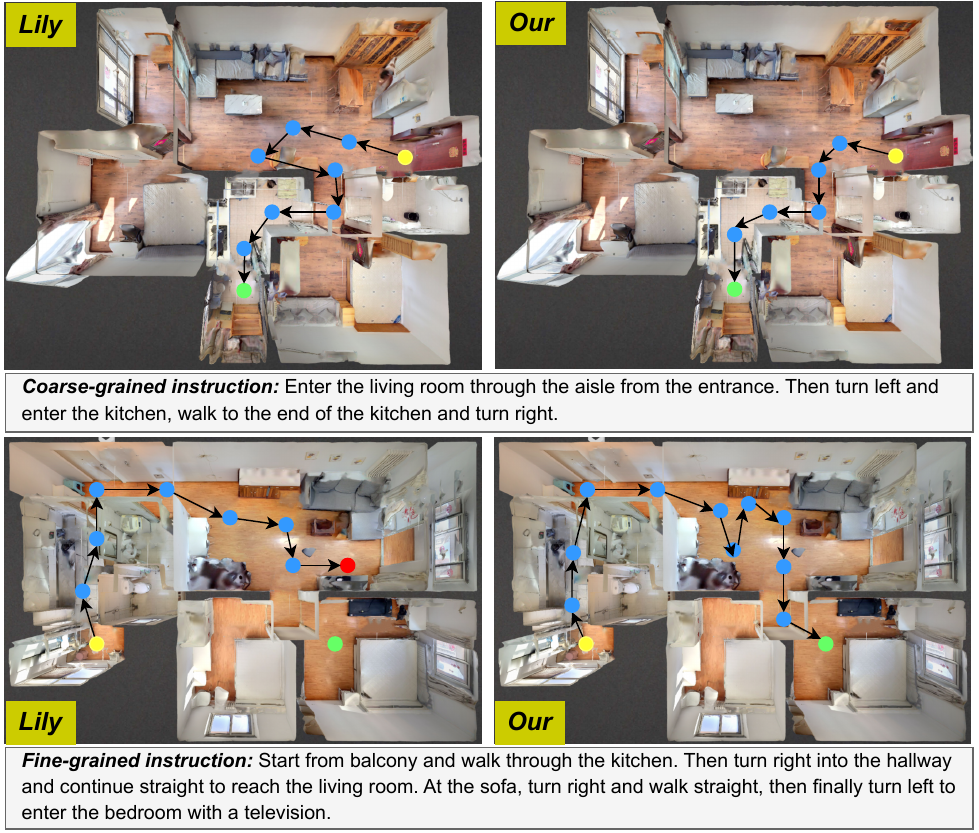}
\caption{The model's performance with instructions of different granularities on real scenes. Yellow and green circles denote the start and target locations, respectively, and the red circles represent incorrect endpoints.}
\label{navigation visiual}
\end{figure}

In Fig. \ref{navigation visiual}, we visualize the navigation paths of our model and Lily when following instructions for different granularities. Lily is trained using instructions generated from the R2R template, while our model is trained using the R2R* instructions. Although Lily successfully reached the destination for coarse-grained navigation instructions, its path was more circuitous compared to our model, further validating that precise instructions can improve navigation efficiency. For fine-grained navigation instructions, Lily failed to reach the destination. While our model’s path was somewhat winding, it arrived at its destination. It demonstrates the positive impact of instruction richness on model performance.

\subsection{Comparison with State-of-the-Arts}

In the R2R, we compare our method in detail with other advanced methods. Table \ref{R2R} shows that our model performance has improved compared to VLNbert and Airbert pre-trained using web image capture pairs. The improvement is attributed to the use of path instruction pairs constructed from YouTube videos with rich real-world navigation experiences, which significantly enhances the navigation performance of the agent. Further, compared with Lily, our model's Success Rate (SR) on the seen validation set increased from 79.31 to 81.23, and on the unseen validation set, the SR improved from 70.00 to 71.10. These enhancements validate the advantages of our high-quality navigation instructions. More precise instructions, finely aligned with visual observations, facilitate a deeper understanding of the environment, improving the agent's adaptability across different environments.

\begin{table}[!t]
{
\resizebox{1.0\linewidth}{!}{
    \begin{tabular}{c|cccc|cccc}
    \hline
    \multirow{2}{*}{Methods} & \multicolumn{4}{c|}{Val-Seen} & \multicolumn{4}{c}{Val-Unseen} \\ \cline{2-9} 
                         & TL    & NE ↓   & SR ↑   & SPL ↑  & TL    & NE ↓   & SR ↑    & SPL ↑  \\ \hline
Follower \cite{fried2018speaker}                 & 10.40  & 3.68 & 65.10 & 0.62 & 9.57  & 5.20  & 52.36  & 0.49  \\ 
Speaker \cite{fried2018speaker}                 & 11.19  & 3.80 & 60.69 & 0.56 & 10.71  & 4.25  & 54.66  & 0.49  \\ 
Speaker-Follower \cite{fried2018speaker}        & 10.69  & \textbf{2.72} & 74.22 & 0.70 & 10.10  & 3.32  & 67.90  & 0.63  \\ 
ProbES \cite{Prompt}                  & -      & -    & -     & -    & 9.50  & 4.05  & 60.28  & 0.56  \\ 
VLNbert \cite{Improving}                 & 10.28  & 3.73 & 70.20 & 0.66 & 9.60  & 4.10  & 59.26  & 0.55  \\ 
Airbert \cite{guhur2021airbert}                 & 10.59  & 3.21 & 73.85 & 0.69 & 10.03  & 3.24  & 68.67  & 0.63  \\ 
Lily \cite{lin2023learning}                    & 10.21  & 2.89 & 79.31 & 0.76 & 10.03  & 3.19  & 70.00  & 0.65  \\ \hline
Ours      & 10.14    & 2.78    & \textbf{81.23}   & \textbf{0.78}   & 9.80  & \textbf{3.13}  & \textbf{71.10}   & \textbf{0.66}      \\ \hline
    \end{tabular}
}
}
\centering
\caption{Comparison with mainstream methods on R2R dataset.}
\label{R2R}
\end{table}

\subsection{Zero/One-Shot Navigation}

\textbf{One-Shot task.} We hypothesize that pre-training with high-quality and diverse path-instruction pairs can achieve exceptional performance with minimal training data. We allow the model to learn from a single environment to evaluate this hypothesis. We randomly sample five sets of environments and report the average results. Since fine-tuning VLNbert and Airbert on the R2R relies on candidate paths extracted from an existing model (EnvDrop \cite{EnvDrop}), candidate paths are sampled as the shortest path between two random positions to ensure fairness. Table \ref{one-shot} presents the comparison results of our method with other methods on the R2R. Specifically, our agent outperforms existing pre-training methods in Val-Seen and Val-Unseen environments. The experimental results validate our hypothesis.

\begin{table}[!t]
\centering
\scalebox{0.85}{
\begin{tabular}{c|c|c}
\hline
Methods & Val Seen       & Val Unseen     \\ \hline
VLNbert \cite{Improving} & 45.71          & 22.43          \\ 
Airbert \cite{guhur2021airbert} & 47.88          & 50.00          \\ 
Lily \cite{lin2023learning}   & 49.31          & 50.86          \\ \hline
Ours     & \textbf{50.43} & \textbf{51.25} \\ \hline
\end{tabular}
}
\caption{SR on val-seen and val-unseen splits of R2R. All the agents access only one environment.}
\label{one-shot}
\end{table}

\textbf{Zero-shot task.} Zero-shot navigation ability is essential for agents as it significantly enhances their generalization in unseen environments, reduces reliance on extensive labelled data, and thus lowers training costs and time. To verify the effectiveness of our method, we pre-trained and fine-tuned our model using \#6 from Table \ref{style}, and tested the model's zero-shot navigation ability on the RxR. Table \ref{zero-shot} shows that our model achieves optimal performance on the val-unseen split. It indicates that using high-quality and diverse training resources can help the agent learn useful environmental features and broader navigation knowledge, significantly improving the agent's generalization ability.

\begin{table}[!t]
\centering
\scalebox{0.85}{
\begin{tabular}{c|cccc}
\hline
\multirow{2}{*}{Methods} & \multicolumn{4}{c}{RxR Val-Unseen} \\ \cline{2-5} 
                         & TL     & NE ↓    & SR  ↑    & SPL  ↑  \\ \hline
DUET \cite{DUET}                    & -      & -      & 23.05   & 0.18   \\ 
Lily \cite{lin2023learning}      & -      & -      & 27.20   & 0.20   \\ \hline
Ours                      & 8.13   & \textbf{3.02} & \textbf{29.35} & \textbf{0.22}   \\ \hline
\end{tabular}
}
\caption{Zero-shot navigation performance of agents on the RxR benchmark.}
\label{zero-shot}
\end{table}

\section{CONCLUSIONS And  Limitations}

In our paper, we propose \textit{InstruGen}, a novel Vision-and-Language Navigation (VLN) path-instruction pairs generation paradigm. Leveraging the powerful visual understanding and generation abilities of large multimodal models (LMMs), \textit{InstruGen} enables the automatic generation of high-quality, multi-granularity navigation instructions. We also contribute a real-world VLN dataset based on YouTube house tour videos. To effectively reduce hallucinations of LMMs, we design a multi-stage verification mechanism to ensure the accuracy and consistency of the instructions. Experimental results show that agents trained with path-instruction pairs generated by \textit{InstruGen} perform excellently on the R2R and RXR benchmarks, particularly demonstrating outstanding generalization abilities in unseen environments. It validates the critical role of high-quality navigation training resources in enhancing agent performance and showcases the potential of large multimodal models in VLN training resource generation. However, our method also has some limitations, such as sampling discrete navigation trajectories similar to R2R, which may be restricted in continuous navigation scenes. We will address this issue in our next work.

\newpage

\bibliographystyle{IEEEtran}
\bibliography{example}

\end{document}